# An Empirical Evaluation of Prompting Strategies for Large Language Models in Zero-Shot Clinical Natural Language Processing

**Sonish Sivarajkumar[1], Mark Kelley[2], Alyssa Samolyk-Mazzanti[2], Shyam Visweswaran, MD, PhD[1,3], Yanshan Wang, PhD[1,2,3*]**

[1]Intelligent Systems Program, University of Pittsburgh, Pittsburgh, PA
[2]Department of Health Information Management, University of Pittsburgh, Pittsburgh, PA
[3]Department of Biomedical Informatics, University of Pittsburgh, Pittsburgh, PA

**Abstract**
*Large language models (LLMs) have shown remarkable capabilities in Natural Language Processing (NLP), especially in domains where labeled data is scarce or expensive, such as clinical domain. However, to unlock the clinical knowledge hidden in these LLMs, we need to design effective prompts that can guide them to perform specific clinical NLP tasks without any task-specific training data. This is known as in-context learning, which is an art and science that requires understanding the strengths and weaknesses of different LLMs and prompt engineering approaches. In this paper, we present a comprehensive and systematic experimental study on prompt engineering for five clinical NLP tasks: Clinical Sense Disambiguation, Biomedical Evidence Extraction, Coreference Resolution, Medication Status Extraction, and Medication Attribute Extraction. We assessed the prompts proposed in recent literature, including simple prefix, simple cloze, chain of thought, and anticipatory prompts, and introduced two new types of prompts, namely heuristic prompting and ensemble prompting. We evaluated the performance of these prompts on three state-of-the-art LLMs: GPT-3.5, BARD, and LLAMA2. We also contrasted zero-shot prompting with few-shot prompting, and provide novel insights and guidelines for prompt engineering for LLMs in clinical NLP. To the best of our knowledge, this is one of the first works on the empirical evaluation of different prompt engineering approaches for clinical NLP in this era of generative AI, and we hope that it will inspire and inform future research in this area.*

## 1. Introduction

Clinical information extraction (IE) is the task of identifying and extracting relevant information from clinical narratives, such as clinical notes, radiology reports, or pathology reports. Clinical IE has many applications in healthcare, such as improving diagnosis, treatment, and decision making, facilitating clinical research, and enhancing patient care(1, 2). However, clinical IE faces several challenges, such as the scarcity and heterogeneity of annotated data, the complexity and variability of clinical language, and the need for domain knowledge and expertise.

Zero-shot IE is a promising paradigm that aims to overcome these challenges by leveraging large pre-trained language models (LMs) that can perform IE tasks without any task-specific training data(3). In-context learning is a framework for zero-shot and few-shot learning, where a large pre-trained LM takes a context, and directly decodes the output without any retraining or fine-tuning(4). In-context learning relies on prompt engineering, which is the process of crafting informative and contextually relevant instructions or queries as inputs to language models to guide their output for specific tasks(5). The utility of prompt engineering lies in its ability to leverage the powerful capabilities of LLMS, such as GPT-3.5(6), LLAMA-2(7), even in scenarios where limited or no task-specific training data is available. In clinical NLP, where labeled datasets tend to be scarce, expensive and time-consuming to create, splintered across institutions, and constrained by data use agreements, prompt engineering becomes even more crucial to unlock the potential of state-of-the-art language models for clinical NLP tasks.

While prompt engineering has been widely explored for general NLP tasks, its application and impact in clinical NLP remain relatively unexplored. Most of the existing literature on prompt engineering in the healthcare domain focuses on biomedical NLP tasks, rather than clinical NLP tasks that involve processing real-world clinical notes. For instance,

---
[*] Corresponding author: yanshan.wang@pitt.edu

Chen et al.(8) used a fixed template as the prompt to measure the performance of LLMs on biomedical NLP tasks, but did not investigate different kinds of prompting methods. Wang et al.(9) gave a comprehensive survey of prompt engineering for healthcare NLP applications such as question-answering systems, text summarization, and machine translation. However, they did not compare and evaluate different types of prompts for specific clinical NLP tasks and how the performance varies across different LLMs. There is a lack of systematic and comprehensive studies on how to engineer prompts for clinical NLP tasks, and the existing literature predominantly focuses on general NLP problems. This creates a notable gap in the research, warranting a dedicated investigation into the design and development of effective prompts specifically for clinical NLP. Currently, researchers in the field lack a comprehensive understanding of the types of prompts that exist, their relative effectiveness, and the challenges associated with their implementation in clinical settings.

The main research question and objectives of our study are to investigate how to engineer prompts for clinical NLP tasks, identify best practices, and address the challenges in this emerging field. By doing so, we aim to propose a guideline for future prompt-based clinical NLP studies. In this work, we present a comprehensive empirical evaluation study on prompt engineering for five diverse clinical NLP tasks, namely Clinical Sense Disambiguation, Biomedical Evidence Extraction, Coreference Resolution, Medication Status Extraction, and Medication Attribute Extraction (10, 11). By systematically evaluating different types of prompts proposed in recent literature, including prefix (12), cloze(13), chain of thought (14), and anticipatory prompts(15), we gain insights into their performance and suitability for each task. We also introduce two new types of prompting approaches: 1) Heuristic prompts and 2) Ensemble. The rationale behind these novel prompts is to leverage the existing knowledge and expertise in rule-based NLP, which has been prominent and has shown significant results in the clinical (16) We hypothesize that heuristic prompts, which are based on rules derived from domain knowledge and linguistic patterns, can capture the salient features and constraints of the clinical IE tasks. We also conjecture that ensemble prompts, which are composed of multiple types of prompts, can benefit from the complementary strengths and mitigate the weaknesses of each individual prompt.

One of the key aspects of prompt engineering is the number of examples or shots that are provided to the model along with the prompt. Few-shot prompting is a technique that provides the model with a few examples of input-output pairs, while zero-shot prompting does not provide any examples(3, 17). By contrasting these strategies, we aim to shed light on the most efficient and effective ways to leverage prompt engineering in clinical NLP. Finally, we propose a prompt engineering framework to build and deploy zero-shot NLP models for clinical domain. Our study covers three state-of-the-art language models, including GPT-3.5, BARD (Chatbot by Google based on PALM-2), and LLAMA 2, to assess the generalizability of the findings across various models. Our work yields novel insights and guidelines for prompt engineering specifically for clinical NLP tasks.

## 2. Methods
### 2.1 Tasks
We selected five distinct clinical NLP tasks representing diverse categories of natural language understanding: Clinical Sense Disambiguation (text classification)(18), Biomedical Evidence Extraction (NER)(19), Coreference Resolution(20), Medication Status Extraction (NER + classification)(21), and Medication Attribute Extraction (NER + RE)(22). Table 1 provides a succinct overview of each task, an example scenario, and the corresponding prompt type employed for each task.

**Table 1:** Task Descriptions

| Task | NLP Task Category | Description | Dataset Example | Example Prompt |
|---|---|---|---|---|
| Clinical Sense Disambiguation | Text Classification | This task involves identifying the correct meaning of clinical abbreviations within a given context. | The abbreviation "CR" can refer to "cardiac resuscitation" or "computed radiography". | What is the meaning of the abbreviation CR in the context of cardiology? |

| Task | Sub-task | Description | Example | Prompt |
|---|---|---|---|---|
| Biomedical Evidence Extraction | Text Extraction | In this task, interventions are extracted from biomedical abstracts. | Identifying panic, avoidance, and agoraphobia (psychological interventions) | Identify the psychological interventions in the given text? |
| Coreference Resolution | Coreference Resolution | The goal here is to identify all mentions in clinical text that refer to the same entity. | Resolving references to "the patient" or "the study" within a clinical trial report. | Identify the antecedent for the patient in the clinical note. |
| Medication Status Extraction | Named Entity Recognition + classification | This task involves identifying whether a medication is currently being taken, not taken, or unknown. | Identifying that a patient is currently taking insulin for diabetes. | What is the current status of [medication name] in the treatment of [medical condition]? |
| Medication Attribute Extraction | Named Entity Recognition + Relation Extraction | The objective here is to identify specific attributes of a medication, such as dosage and frequency. | Identifying dosage, frequency, and route of a medication for a patient. | What is the recommended dosage of [medication name] for [patient population] and how often? |

## 2.2 Datasets and Evaluation

The prompts were evaluated on three language models, GPT-3.5, BARD(which is a chatbot based on PALM-2), LLAMA2, under both zero-shot and few-shot prompting conditions, employing precise experimental settings and parameters. To simplify the evaluation process and facilitate clear comparisons, we adopted accuracy as the sole evaluation metric for all tasks. Accuracy is defined as the proportion of correct outputs generated by the language model for each task, using a resolver that maps the output to the label space. Table 3 shows the datasets and evaluation metrics for each clinical NLP task. The datasets are:

- Clinical Abbreviation Sense Inventories(CASI): This is a dataset of clinical abbreviations, senses, and instances(23).It contains 41 acronyms from 18,164 notes, along with their expanded forms and contexts. We used a randomly sampled subset from this dataset for clinical sense disambiguation, coreference resolution, medication status extraction, and medication attribute extraction tasks(Table 2).
- EBM-NLP: This is a dataset of evidence-based medicine (EBM) annotations for NLP (24). It contains 187 abstracts and 20 annotated abstracts, with interventions extracted from the text. We used this dataset for biomedical evidence extraction task.

**Table 2:** Evaluation Datasets, and Samples for Different Tasks

| Task | Dataset | Samples | Metric |
|---|---|---|---|
| Clinical Sense Disambiguation | CASI | 11 acronyms from 55 notes | Accuracy: proportion of correct senses extracted |
| Biomedical Evidence Extraction | EBM-NLP | 187 abstracts and 20 annotated abstracts | Accuracy: proportion of correct entities or relations extracted |
| Coreference Resolution | CASI | 105 annotated examples | Accuracy: proportion of correct coreference resolutions |

| Medication Status Extraction | CASI | 105 annotated examples with 340 medication-status pairs | Accuracy: proportion of correct status extracted for each medication |
| --- | --- | --- | --- |
| Medication Attribute Extraction | CASI | 105 annotated examples with 313 medications and 533 attributes | Accuracy: proportion of correct attribute extracted for each medication |

## 2.2 Prompt Creation Process

We followed a rigorous process to create suitable prompts for each task. These prompts were carefully crafted to match the specific context and objectives of each task. There is no established method for prompt design and selection as of now. Therefore, we adopted an iterative approach where prompts, which are created by healthcare experts, go through a verification and improvement process in an iterative cycle, which involved design, experimentation, and evaluation, as depicted in Figure 2.

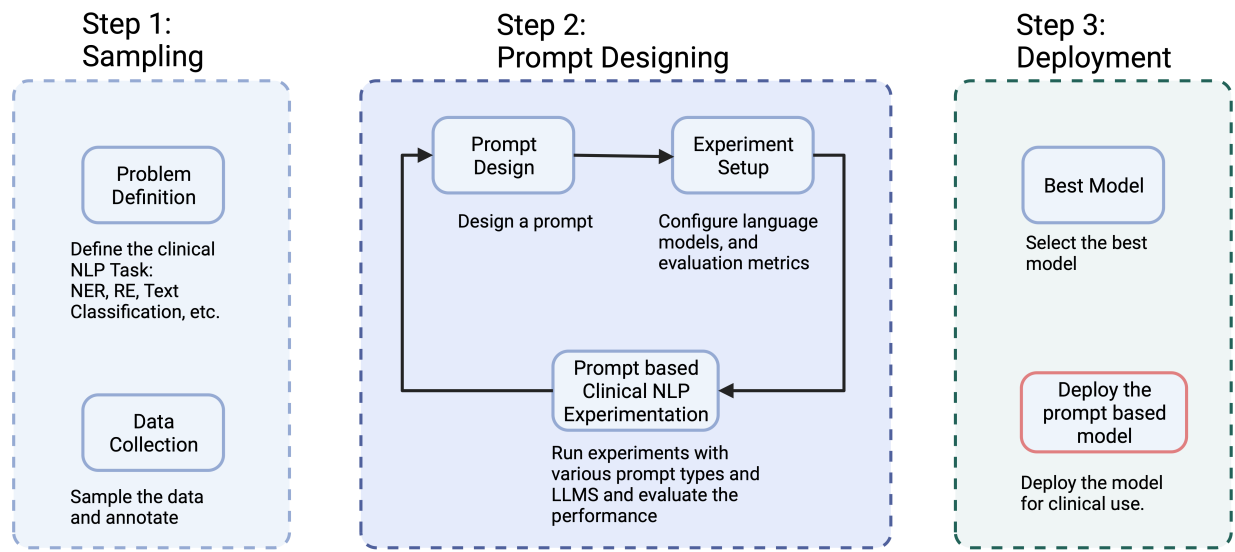

**Figure 2:** Iterative Prompt Design Process: A schematic diagram of the iterative prompt creation process for clinical NLP tasks. The process consists of three steps: sampling, prompt designing, and deployment. The sampling step involves defining the task and collecting data and annotations. The prompt designing step involves creating and refining prompts using different types and language models. The deployment step involves selection of the best model and deploying the model for clinical use.

Figure 2 illustrates the three main steps of our prompt creation process: sampling, prompt designing, and deployment. In the sampling step(step 1), we defined the clinical NLP task (e.g., NER, RE, text classification, etc.) and collected a sample of data and annotations as an evaluation for the task. In the prompt designing step(step 2), we designed a prompt for the task using one of the prompt types (e.g., simple prefix prompt, simple cloze prompt, heuristic prompt, chain of thought prompt, question prompt, anticipatory prompt, etc.). We also optionally performed few-shot prompting by providing some examples along with the prompt. We then configured the LLMs and the evaluation metrics for the experiment setup. We ran experiments with various prompt types and language models and evaluated their performance on the task. Based on the results, we refined or modified the prompt design until we achieved satisfactory performance or reached a limit. In the deployment step(step 3), we the best prompt-based models based on their performance metrics and deployed the model for the corresponding task.

## 2.3 Prompt Engineering Techniques

Prompt engineering is the process of designing and creating prompts that elicit desired responses from language models. Prompts can be categorized into different types based on their structure, function, and complexity.

Each prompt consists of a natural language query that is designed to elicit a specific response from the pre-trained language model. The prompts are categorized into seven types, as illustrated in Figure 3. Prefix prompts are the simplest type of prompts, which prepend a word or phrase indicating type/format/tone of response, for control and relevance. Cloze prompts are based on the idea of fill-in-the-blank exercises, which create a masked token in the input text and ask the language model to predict the missing word or phrase. Anticipatory prompts are the prompts anticipating the next question/command based on experience or knowledge, guiding conversation.

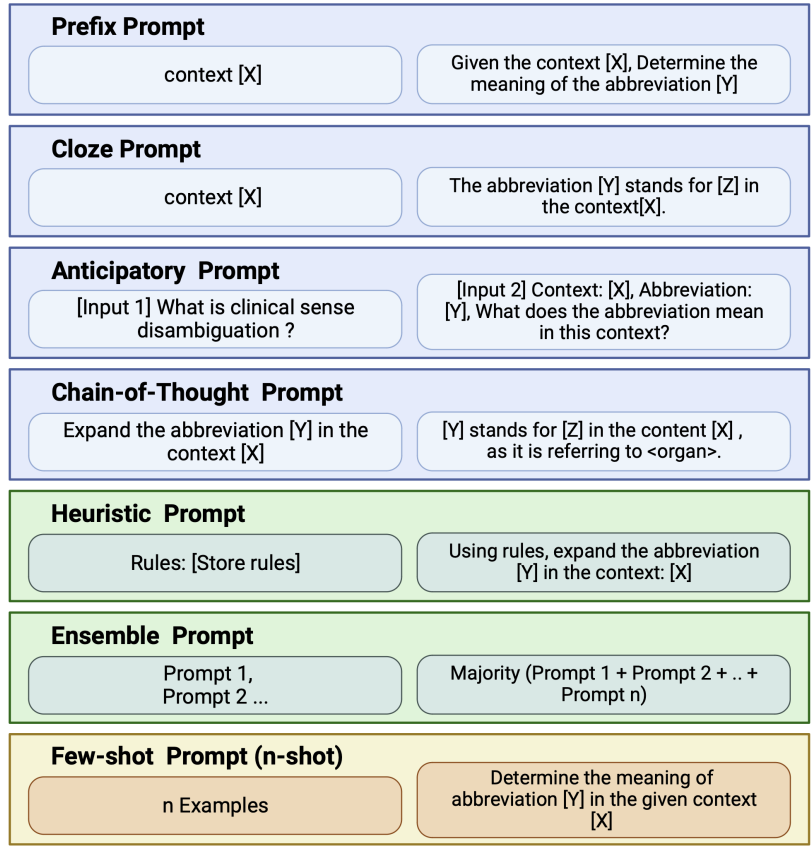

**Figure 3:** Types of prompts: Examples of six types of prompts that we used to query the pre-trained language model for different clinical IE tasks. [X]: context , [Y]: Abbreviation, [Z]: Expanded Form.(All prompts have been included in Appendix A.)

In addition to the existing types of prompts, we also designed two new novel prompts: heuristic prompts and ensemble prompts. **Heuristic prompts** are rule-based prompts that break down complex queries into smaller parts for comprehensive answers. They use a set of predefined rules to guide the language model to expand the abbreviations in a given context. For example, a heuristic prompt can use the rule that an abbreviation is usually capitalized, followed by a period, and preceded by an article or a noun. Unlike chain of thought prompts which explain the reasoning or logic behind the output, heuristic prompts use a set of predefined rules to guide the language model to perform a task.

**Ensemble prompts** are prompts that combine multiple prompts using majority voting for aggregated outputs. They use different types of prompts to generate multiple outputs for the same input, and then select the most common output as the final answer. For example, an ensemble prompt can use three different prefix prompts or combination of all other types of prompts to generate three possible expansions for an abbreviation, and then choose the one that appears most frequently. For simplicity, we combined all the five different prompt outputs using a majority voting approach.The intuition behind an ensemble prompt is that by combining multiple types of prompts, we can leverage the strengths and compensate for the weaknesses of each individual prompt. For example, some prompts may be more effective for certain tasks or certain models than others, or some prompts may be more robust to noise or ambiguity than others. By using majority voting, we can select the output that is most likely to be correct or coherent among the outputs generated by different types of prompts.

## 3. Results

In this section, we present the results of our experiments on prompt engineering for zero-shot clinical information extraction. We evaluated various prompt types across five clinical NLP tasks, aiming to understand how different prompts influence the accuracy of different language models. We also compared zero-shot and few-shot prompting strategies, exploring how the addition of context affects the model performance. Furthermore, we tested an ensemble approach that combines the outputs of different prompt types using majority voting. Finally, we analyzed the impact of different language models on the task performance and observed some interesting patterns. Table 3 illustrates that different prompt types have different levels of effectiveness for different tasks and language models. We can also observe some general trends across the tasks and models.

### 3.1 Prompt Optimization and Evaluation

*Table 3: Performance Comparison of Different Prompt Types and Language Models*

| Task | Language Model | Simple Prefix | Simple Cloze | Anticip-atory | Heuristic | Chain of Thought | Ensemble | Few-shot |
|---|---|---|---|---|---|---|---|---|
| Clinical Sense Disambiguation | GPT 3.5 | 0.88 | 0.86 | 0.88 | **0.96** | 0.9 | 0.9 | 0.82 |
| | BARD | **0.76** | 0.68 | 0.71 | 0.75 | 0.72 | 0.71 | 0.67 |
| | LLAMA2 | **0.88** | 0.76 | 0.82 | 0.82 | 0.78 | 0.82 | 0.78 |
| Biomedical Evidence extraction | GPT 3.5 | 0.92 | 0.82 | 0.88 | 0.94 | 0.94 | 0.88 | **0.96** |
| | BARD | 0.89 | 0.89 | **0.91** | 0.9 | **0.91** | 0.9 | 0.88 |
| | LLAMA2 | 0.85 | **0.88** | 0.87 | **0.88** | 0.87 | 0.88 | 0.86 |
| Coreference Resolution | GPT 3.5 | 0.78 | 0.6 | 0.74 | **0.94** | **0.94** | 0.74 | 0.74 |
| | BARD | 0.69 | **0.81** | 0.73 | 0.67 | 0.71 | 0.69 | 0.7 |
| | LLAMA2 | **0.8** | 0.64 | 0.74 | 0.76 | **0.8** | 0.78 | 0.68 |
| Medical Status Extraction | GPT 3.5 | **0.76** | 0.72 | 0.75 | 0.74 | 0.73 | 0.75 | 0.72 |
| | BARD | **0.67** | 0.51 | 0.65 | 0.55 | 0.59 | 0.58 | 0.55 |
| | LLAMA2 | 0.58 | 0.48 | 0.52 | **0.64** | 0.52 | 0.58 | 0.42 |
| Medical Attribute Extraction | GPT 3.5 | 0.88 | 0.84 | 0.9 | **0.96** | **0.96** | 0.9 | **0.96** |
| | BARD | 0.68 | 0.72 | **0.88** | 0.7 | 0.74 | 0.76 | **0.88** |
| | LLAMA2 | 0.6 | 0.66 | 0.58 | 0.66 | **0.72** | 0.64 | 0.6 |

For clinical sense disambiguation, the heuristic and prefix prompts consistently achieved the highest performance across all language models. This indicates that these prompts were able to provide enough guidance and context for the model to disambiguate the word in the clinical note. Among the language models, GPT 3.5 excelled with this prompt type, achieving an accuracy of 0.96. For biomedical evidence extraction, the heuristic and chain-of-thought prompts excelled across all language models in zero-shot setting. This indicates that these prompt types were able to provide enough information and constraints for the model to extract the evidence from the clinical note. GPT 3.5 achieved an accuracy of 0.94 with these prompt types, which was higher than any other model or prompt type combination. For coreference resolution, the chain of thought prompt type performed best among all prompt types with two language models-GPT3.5 and LLAMA2. This indicates that this prompt type was able to provide enough structure and logic for the model to resolve the coreference in the clinical note. GPT 3.5 displayed high accuracy with this prompt type, achieving an accuracy of 0.94. For medication status extraction, simple prefix and heuristic prompts yielded good results across all language models. These prompt types were able to provide enough introduction or rules for the model to extract the status of the medication in relation to the patient or condition. GPT 3.5 excelled with these prompt types, achieving an accuracy of 0.76 and 0.74 respectively. For medication attribute extraction, we found that the chain of thought and heuristic prompts were effective across all language models. These prompt types were able to provide enough reasoning or rules for the model to extract and label the attributes of medications from clinical notes. Anticipatory prompts, however, had the best accuracy for BARD among all the prompts. GPT-3.5 achieved an accuracy of 0.96 with these prompt types, which was higher than any other model or prompt type combination.

Thus, we can see that task-specific prompt tailoring is crucial for achieving high accuracy. Different tasks require different levels of information and constraints to guide the language model to produce the desired output. The experiments show that heuristic, prefix and chain of thought prompts are generally very effective for guiding the language model to produce clear and unambiguous outputs. We have also shown that GPT 3.5 is a superior and versatile language model that can handle various clinical NLP tasks in zero-shot settings, outperforming other models in most cases.

### 3.2 Zero-shot vs Few-shot Prompting

We compared the performance of zero-shot prompting and few-shot prompting strategies for each clinical NLP task. We used the same prompt types and language models as in the previous experiments, but added some context to the input in the form of examples or explanations. We used two examples or explanations for each task (2-shot), depending on the complexity and variability of the task. Table 3 shows that few-shot prompting consistently improved the accuracy of all combinations for all tasks, except for clinical sense disambiguation and medication attribute extraction, where some zero-shot prompt types performed better. We also observed some general trends across the tasks and models.

We found that few-shot prompting enhanced accuracy by providing limited context that aided complex scenario understanding. The improvement was more pronounced compared to simple cloze prompts, which had lower accuracy in most of the tasks. We also found that some zero-shot prompt types were very effective for certain tasks, even outperforming few-shot prompting. These prompt types used a rule-based or reasoning approach to generate sentences that contained definitions or examples of the target words or concepts, which helped the language model to understand and match the context. For example, heuristic prompts achieved higher accuracy than few-shot prompting for clinical sense disambiguation and medication attribute extraction, while chain of thought prompts achieved higher accuracy than few-shot prompting for coreference resolution and medication attribute extraction.

From these results, we can infer that LLMs can be effectively used for clinical NLP in a no-data scenario, where we don't have much publicly available datasets, by using appropriate zero-shot prompt types that guide the language model to produce clear and unambiguous outputs. However, few-shot prompting can also improve the performance of LLMs by providing some context that helps the language model to handle complex scenarios.

### 3.3 Other Observations

*Ensemble Approaches*

We experimented with an ensemble approach by combining outputs from multiple prompts using majority voting. The ensemble approach was not the best performing strategy for any of the tasks, but it was better than the low performing prompts. The ensemble approach was able to benefit from the diversity and complementarity of different prompt types, and avoid some of the pitfalls of individual prompts. For example, for clinical sense disambiguation, the ensemble approach achieved an accuracy of 0.9 with GPT 3.5, which was the second best performing prompt type. Similarly, for medical attribute extraction, the ensemble approach achieved an accuracy of 0.9 with GPT 3.5 and 0.76 with BARD, which were close to the best single prompt type (anticipatory). However, the ensemble approach also suffered from some drawbacks, such as inconsistency and noise. For tasks that required more specific or consistent outputs, such as coreference resolution, the ensemble approach did not improve the accuracy over the best single prompt type, and sometimes even decreased it. This suggests that the ensemble approach may introduce ambiguity for tasks that require more precise or coherent outputs.

*Impact of Language Models*

We observed variations in performance among different language models(Table 3). We found that GPT 3.5 generally outperformed BARD and LLAMA2 on most tasks. This suggests that GPT 3.5 has a better generalization ability and can handle a variety of clinical NLP tasks with different prompt types. However, BARD and LLAMA2 also showed some advantages over GPT 3.5 on certain tasks and prompt types. For example, BARD achieved the highest accuracy of 0.81 with simple cloze prompts and LLAMA2 achieved the highest accuracy of 0.8 with simple prefix prompts for coreference resolution. This indicates that BARD and LLAMA2 may have some domain-specific knowledge that can benefit certain clinical NLP tasks for specific prompt types.

*Persona Patterns*
Persona patterns are a way of asking the language model to act like a persona or a system that is relevant to the task or domain. For example, one can ask the language model to "act as a clinical NLP expert". This can help the language model to generate outputs that are more appropriate and consistent with the persona or system. For example, one can use the following prompt for clinical sense disambiguation:
Act as a clinical NLP expert. Disambiguate the word "cold" in the following sentence: "She had a cold for three days."
We experimented with persona patterns for different tasks and language models, and found that they can improve the accuracy and quality of the outputs. Persona patterns can help the language model to focus on the relevant information and constraints for the task, and avoid generating outputs that are irrelevant or contradictory to the persona or system.

*Randomness in Output:*
Most LLMs do not produce the output in the same format every time. There is inherent randomness in the outputs the language models produce. Hence, the prompts need to be specific in the way they are done for the task. Prompts are powerful when they are specific and if we use them in the right way.
Randomness in output can be beneficial or detrimental for different tasks and scenarios. In clinical domain, randomness can introduce noise and errors in the outputs, which can make them less accurate and reliable for the users. For example, for tasks that involve extracting factual information, such as biomedical evidence extraction, medical status extraction, etc., randomness can cause the language model to produce outputs that are inconsistent or contradictory with the input or context.

### 3.4 Guidelines and Suggestions for Optimal Prompt Selection
In this section, we synthesize the main findings from our experiments and offer some practical advice for prompt engineering for zero-shot and few-shot clinical information extraction. We propose the following steps for selecting optimal prompts for different tasks and scenarios:
Define the goal and scope of the prompt.

The first step is to identify the type of the clinical NLP task, which can be broadly categorized into three types: 1) Classification, 2) Extraction, and 3) Resolution. Classification tasks involve assigning a label or category to a word, phrase, or sentence in a clinical note, such as clinical sense disambiguation or medication status extraction. Extraction tasks involve identifying and extracting relevant information from a clinical note, such as biomedical evidence extraction or medication attribute extraction. Resolution tasks involve linking or matching entities or concepts in a clinical note, such as coreference resolution.

The second step is to choose the prompt type that is most suitable for the task type. We found that different prompt types have different strengths and weaknesses for different task types, depending on the level of information and constraints they provide to the language model. Table 4a summarizes our findings and recommendations for optimal prompt selection for each task type.

**Table 4a:** Optimal prompt types for different clinical NLP task types

| Task Type | Prompt Type |
|---|---|
| Classification | Heuristic or Prefix |
| Extraction | Heuristic of Chain-of-thought |
| Resolution | Chain-of-thought |

**Table 4b:** Optimal language models for different prompt types

| Prompt Type | Language Model |
|---|---|
| Heuristic | GPT-3.5 |
| Prefix | GPT-3.5 or LLAMA2 |
| Cloze | BARD or LLAMA2 |
| Chain of Thought | GPT-3.5 |
| Anticipatory | BARD |

The third step is to choose the language model that is most compatible with the chosen prompt type. We found that different language models have different capabilities and limitations for different prompt types, depending on their generalization ability and domain-specific knowledge. Table 4b summarizes our findings and recommendations for optimal language model selection for each prompt type.

The fourth step is to evaluate the performance of the chosen prompt type and language model combination on the clinical NLP task, using appropriate metrics such as accuracy, precision, recall, or F1-score. If the performance is satisfactory, then the prompt engineering process is complete. If not, then the process can be repeated by choosing a different prompt type or language model, or by modifying the existing prompt to improve its effectiveness.

**4. Discussion**

In this paper, we have presented a novel approach to zero-shot and few-shot clinical information extraction using prompt engineering. We have evaluated various prompt types across five clinical NLP tasks: clinical sense disambiguation, biomedical evidence extraction, coreference resolution, medical status extraction, and medical attribute extraction. We have also compared the performance of different language models: GPT 3.5, BARD, and LLAMA2. Our main findings are as follows:

(1) Task-specific prompt tailoring is crucial for achieving high accuracy. Different tasks require different levels of information and constraints to guide the language model to produce the desired output. Therefore, it is important to design prompts that are relevant and specific to the task at hand, and avoid using generic or vague prompts that may confuse the model or lead to erroneous outputs.

(2) Heuristic prompts are generally very effective for guiding the language model to produce clear and unambiguous outputs. These prompts use a rule-based approach to generate sentences that contain definitions or examples of the target words or concepts, which help the model to understand and match the context. Heuristic prompts are especially useful for tasks that involve disambiguation, extraction, or classification of entities or relations.

(3) Chain of thought prompts are also effective for guiding the language model to produce logical and coherent outputs. These prompts use a multi-step approach to generate sentences that contain a series of questions and answers that resolve the task in the context. Chain of thought prompts are especially useful for tasks that involve reasoning, inference, or coreference resolution.

(4) Few-shot prompting can improve the performance of language models by providing some context that helps the model to handle complex scenarios. Few-shot prompting can be done by adding some examples or explanations to the input, depending on the complexity and variability of the task. Few-shot prompting can enhance accuracy by providing limited context that aids complex scenario understanding. The improvement is more pronounced compared to simple prefix and cloze prompts, which had lower accuracy in most of the tasks.

(5) Ensemble approaches can also improve the performance of language models by combining outputs from multiple prompts using majority voting. Ensemble approaches can leverage the strengths of each prompt type and reduce the errors of individual prompts. Ensemble approaches are especially effective for tasks that require multiple types of information or reasoning, such as biomedical evidence extraction and medical attribute extraction.

**5. Conclusion**

In this paper, we have benchmarked different prompt engineering techniques for zero-shot clinical NLP tasks. We also conceptualized and proposed two new types of prompts, heuristic and ensemble prompts, we have demonstrated that prompt engineering can enable the use of pre-trained language models for various clinical NLP tasks without requiring any fine-tuning or additional data. We have shown that task-specific prompt tailoring, heuristic prompts, chain of thought prompts, few-shot prompting, and ensemble approaches can improve the accuracy and quality of the outputs. We have also shown that GPT 3.5 is very adaptable and precise across all tasks and prompt types, while BARD and LLAMA2 may have some domain-specific advantages for certain tasks and prompt types.

We believe that prompt-based approach has several benefits over existing methods for clinical information extraction. It reduces the cost and time of developing clinical NLP applications, as it does not require any labeled data or fine-tuning. It is flexible and adaptable, as it can be applied to various clinical NLP tasks with different prompt types and language models. It is interpretable and explainable, as it uses natural language prompts that can be easily understood and modified by humans. It is scalable and portable, as it can leverage the power of pre-trained language models that are widely available and accessible.

# 6. Limitations and Future Work:

**Limitations:**

Our approach also has some limitations that we acknowledge in this work. First, it relies on the quality and availability of pre-trained language models, which may vary depending on the domain and task. As language models are rapidly evolving, some parts of prompt engineering discipline may be timeless, while some parts may evolve and adapt over time as different capabilities of models evolve. Second, it requires a lot of experimentation and iteration to optimize prompts for different applications, which may be tedious and time-consuming. We may not have explored all the possible combinations and variations of prompts that could potentially improve the performance of the clinical NLP tasks. Third, the LLMs do not release the details of the dataset which they were trained on. Hence, the high accuracy could be because the models would have already seen the data during training, and not because of the effectiveness of the prompts.

**Future Work:**

We plan to address these challenges and limitations in our future work. We aim to develop more systematic and automated methods for prompt design and evaluation, such as using prompt-tuning or meta-learning techniques. We also aim to incorporate more domain knowledge or external resources into the prompts or the language models, such as using ontologies, knowledge graphs, or databases. We also aim to incorporate more quality control or error correction mechanisms into the prompts or the language models, such as using adversarial examples, confidence scores, or human feedback.

**Author Information**

**Contributions**

SS conceptualized, designed, and organized this study, analyzed the results, and wrote, reviewed, and revised the paper. MK, AS analyzed the results, and wrote, reviewed, and revised the paper. SV wrote, reviewed, and revised the paper. Y.W. conceptualized, designed, and directed this study, wrote, reviewed, and revised the paper.

**Corresponding author**

Correspondence to Yanshan Wang.

**Ethics declarations**

**Competing Interests**

All authors declare no competing interests.

**Data availability**

Not applicable.

**Code availability**

Not applicable.

**Acknowledgements**

This work was supported by the National Institutes of Health under award number U24 TR004111 and R01 LM014306. The content is solely the responsibility of the authors and does not necessarily represent the official views of the National Institutes of Health.

# Supplementary Material

**Appendix A: Prompts for Clinical NLP Tasks**

In the appendix, we present the prompts that we used for each of the five clinical NLP tasks and each of the five prompt types in our experiments(Table A1). The prompts are written in natural language and formatted according to the input specifications of each language model. The prompt types are: Prefix, Cloze, Anticipatory, Heuristic, and Chain of Thought. We used seven types of prompts in our experiments, but two of them were not independent prompt types. Ensemble prompt type was a combination of the outputs from all the other prompt types, using majority voting to decide the final output. Few-shot prompting was a variation of the other prompt types, where we added two examples(2-shot) to the input to provide some context to the model. The clinical NLP tasks are: Clinical Sense Disambiguation, Biomedical Evidence Extraction, Coreference Resolution, Medication Status Extraction, and Medication Attribute Extraction.

*Table A1: Prompts for each task and prompt type combination*

| Task | Prefix | Cloze | Anticipatory | Heuristic | Chain-of-thought |
|---|---|---|---|---|---|
| Clinical Sense Disambiguation | Determine the meaning to the abbreviation <abbreviation> in the clinical note. Clinical note: | The abbreviation <abbreviation> stands for ___ in the clinical note. Clinical Note: | [Input 1:] what is clinical sense disambiguation? [Input 2:] how to determine what a clinical abbreviation means in a given context? [Input 3:] Context: <text>, Abbreviation: <abbreviation>, What does the abbreviation mean in this context? | [Input 1:] First, store these rules in memory: 1. If the terms "hemoglobin," "ekg," or "valve" appear in the text, "ab" likely stands for "atrioventricular", "avr" likely stands for "aortic valve replacement","av" likely stands for "atrioventricular","cvp" likely stands for "central venous pressure","cvs" likely stands for cardiovascular system","la" likely stands for "left atrial", and "sa" likely stands for "sinuatrial" 2. If the text is discussing cardiology, cardiovascular issues, or the hearts electrical system, "ab" likely stands for "atrioventricular", "avr" likely stands for "aortic valve replacement","av" likely stands for "atrioventricular","cvp" likely stands for "central venous pressure","cvs" likely stands for cardiovascular system","la" likely stands for "left atrial", and "sa" likely stands for "sinuatrial" | EXAMPLE: Expand the abbreviation "<abbreviation>" in the clinical note. Clinical Note: ANSWER: "<abbreviation>" stands for "<expanded form>" in the text because the context is referring to cardiology and the heart. QUESTION: Expand the abbreviation "<abbreviation>" in the clinical note. Clinical Note. |

| | | | | 3. If the text is relating to the heart, heart disease, or any anatomy relating to the heart, "ab" likely stands for "atrioventricular", "avr" likely stands for "aortic valve replacement","av" likely stands for "atrioventricular","cvp" likely stands for "central venous pressure","cvs" likely stands for cardiovascular system","la" likely stands for "left atrial", and "sa" likely stands for "sinuatrial"<br>[Input 2:] Using the stored rules, expand the abbreviation "\<abbreviation\>" in the text: "\<text\>" | |
|---|---|---|---|---|---|
| Biomedical Evidence Extraction | Label each token in the clinical note that references an intervention. If the token is an intervention, specify the type of intervention as: 1=surgical, 2=physical, 3=pharmacological,4=educational,5=psychological,6=other,7= control | For each token in the clinical note below using white space tokenization, the token relates to a ___ intervention. Your choices are: surgical, physical, pharmacological, educational, psychological, other type, control, or not an intervention. | [Input 1:] what is medical intervention.<br>[Input 2:] how to identify medical interventions:<br>[Input 2:] What intervention category does each token in the clinical note belong to? Surgical (1), physical (2), pharmacological (3), educational (4), psychological (5), other (6), control (7), or none (0). Use whitespace tokenization. Clinical Note: | [Input 1:] First, store these rules in memory:<br><br>[Input 2:] Rule 1: Review each token and decide if the token can be classified as a "clinical intervention,". Rule 2: If the token is not an intervention, assign "0" to the token, Rule 3: If the token is an intervention, use the following sub-category to classify the type of intervention: 1 = surgical, 2 = physical, 3 = pharmacological, 4 = educational, 5 = psychological, 6 = other, 7 = control.<br><br>[Input 3:] Using the above rules, extract the medical interventions in the clinical note. Clinical Note: | EXAMPLE: Assign each token in the clinical note to an intervention category: surgical (1), physical (2), pharmacological (3), educational (4), psychological (5), other (6), control (7), or none (0). Use whitespace tokenization. Clinical Note: "Report from Pain Study . Patients with opioid use responded better to treatment . Predicted outcome with morphine or vicodin ."<br>ANSWER: Tokenized text using whitespace tokenization: "Report" "from" "Pain" "Study" "." "Patients" "with" "opioid" "use" "responded" "better" "to" "treatment" "." Predicted" "outcome" "with" "morphine" "or" "vicodin" "." A study is educational so "Pain" and "Study" will be assigned "4." A treatment and drug names are pharmacological so "treatment" |
| | | | | | "morphine" and "vicodin" will be assigned "3." QUESTION: Assign each token in the clinical note to an intervention category: surgical (1), physical (2), pharmacological (3), educational (4), psychological (5), other (6), control (7), or none (0). Use whitespace tokenization. Clinical Note: |
| Coreference Resolution | Identify the antecedent for the pronoun <pronoun> in the clinical note. Clinical Note: | The pronoun <pronoun> refers to ___ in the clinical note. Clinical Note: | [Input 1:] what is coreference resolution? [Input 2:] what is an antecedent? [Input 3:] What is the antecedent for the pronoun <pronoun> in the clinical note? Clinical Note: | [Input 1:] First, store these rules in memory.provided rules: identify the pronoun mentioned in the task, locate the pronoun in the text, determine the noun or noun phrase that the pronoun refers to by considering the context.> [Input 2:] Identify the pronoun <pronoun> in the clinical text. [Input 3:]identify the reference for the pronoun in the clinical note. Clinical Note: | EXAMPLE" Identify the antecedent the pronoun "this" refers to in the clinical note. Clinical Note: "5. Zosyn. This was stopped today, _%#MMDD2003#%_. 6. Multivitamin one tablet p.o. q.d. 7. Lasix 80 mg p.o. q.d. 8. Paxil 40 mg p.o. q.d. 9. Bactrim SS one tablet p.o. Mondays and Thursdays. 10. Ranitidine 150 mg p.o. b.i.d. 11. Metoprolol 25 mg p.o. b.i.d. 12. Colace 100 mg p.o. b.i.d." ANSWER: The pronoun "this" is referring to "Zosyn" because it is the most recent noun or noun phrase mentioned previously. QUESTION: identify the antecedent the pronoun "[pronoun]" refers to in the clinical note. Clinical Note: |
| Medication Status Extraction | In the clinical note, extract the medication and it's current status as active, discontinued, or neither. Clinical note: | The active medications in the text are ___, the discontinued medications are ___, and the medications that are neither active nor discontinued in the clinical note are ___. Clinical Note: | [Input 1:] how to identify contextual level information in a text? [Input 2:] For each medication mentioned in the given clinical note, is the current status "active," "discontinued," or "neither?" | [Input 1:] First, store these rules in memory: <10 rules> | EXAMPLE: Label any medications in the clinical note as active, discontinued, or neither. Text: "2. Prior major leg infection. 3. Penicillin allergy. PLANS: Add Fortaz to vancomycin and DC clindamycin. Will also give one dose of tobramycin. If we do another I&D or dressing change, it would be of value to do swab culture of the wound to see what is colonizing it superficially." |

| | | | | | [Input 2:] Given the clinical note, extract the medication and its current status as active, discontinued, or neither, using the rules mentioned above. Clinical Note: | ANSWER: 1. Penicillin: Neither (doesn't specifically mention active or discontinued, if it's an allergy it's labeled as neither), 2. Fortaz: Active (states it was added to vancomycin, adding indicates it's an active medication), 3. Vancomycin: Active (text states Fortaz was added to Vancomycin, adding indicates it's an active medication), 4. Clindamycin: Discontinued (DC stands for discontinued). QUESTION: Using the stored example, label any medications in the text as active, discontinued, or neither. Text: <text> |
|---|---|---|---|---|---|---|
| Medication Attribute Extraction | Extract the medications mentioned in the clinical note, and any attributes such as route, frequency, dosage, duration, or reason for prescribing if mentioned. Clinical note: | The medications in the clinical note are ___. The name, dosage, frequency, and reason of each medication is ___. Clinical Note: | [Input 1:] how to extract contextual level information in a text such as status about medications? [Input 2:] how to identify specific medication attributes, such as route, frequency, or duration? [Input 3:] In the given clinical note, for each medication mentioned, what are the attributes such as route, frequency, dosage, duration, or reason for prescription? Clinical note: | [Input 1:] Store these rules in memory: <10 rules> [Input 2:] First extract the medications mentioned in the clinical note. Second extract any attributes such as route, frequency, dosage, duration, or reason for prescribing. | EXAMPLE: Identify any medications in the text and list the name and if applicable the dosage, frequency, and reason. Text: "The patient tells me that she was taking 325 mg of aspirin per day for three years for a transient ischemic attack before she had her AVR. Now she is off this and on Coumadin, although the AVR is said to be a tissue valve. She takes no other nonsteroidal anti-inflammatory drugs except occasional Aleve in the past." ANSWER: 1. aspirin 325 mg per day for three years for a transient ischemic attack, 2. Coumadin, 3. nonsteroidal anti-inflammatory drugs, 4. Aleve occasional. QUESTION: Using the stored example, identify any medications |

| | | | | in the clinical note and list the name and if applicable the dosage, frequency, and reason. Clinical Note: |
|---|---|---|---|---|